\documentclass[runningheads]{llncs}

\usepackage{eccv}

\usepackage{eccvabbrv}

\usepackage{booktabs}
\usepackage{multirow}
\usepackage{graphicx}
\usepackage{microtype}
\usepackage{wrapfig} 
\usepackage{xcolor}
\usepackage{colortbl}
\usepackage{booktabs}

\definecolor{headerblue}{RGB}{220, 232, 245}
\definecolor{oursgreen}{RGB}{209, 237, 216}
\definecolor{rowgray}{RGB}{245, 245, 245}

\usepackage{hyperref}

\usepackage{orcidlink}
\newcommand{\methodname}{RapidLiDAR}

\begin{document}

\title{Towards Real-Time and Adaptable LiDAR Scene Completion} 


\author{Azhar Hussian\and
Martin Vossiek\and
Vasileios Belagiannis}

\authorrunning{A.~Hussian et al.}

\institute{Friedrich-Alexander-Universität Erlangen-Nürnberg, Germany\\
\email{\{azhar.hussian,martin.vossiek,vasileios.belagiannis\}@fau.de}}

\maketitle

\begin{abstract}
LiDAR scene completion is a key component of 3D perception in autonomous driving, where the scene must be completed in real time to be usable in downstream tasks. Existing approaches typically follow an initialize-and-refine paradigm, in which a coarse initialization of the scene is first constructed, then refined into complete 3D geometry. Generative models are slower because they iteratively refine random Gaussian noise into the scene, while non-generative methods perturb the partial scene with a fixed noise scale, which limits coverage of large gaps and occluded regions and requires manual recalibration for each new sensor configuration. 
We present RapidLiDAR, a LiDAR scene completion method that treats the initialization itself as a learned, data-driven component. We propose an \textit{adaptive initialization module} that predicts a spatially varying displacement for each partial input point, expanding the partial observations into a coarse scene initialization adapted to the local geometry, without requiring manual noise tuning. To refine this coarse initialization into a complete and coherent scene, we additionally propose a \textit{multi-scale reconstruction module} that further refines point positions by querying multi-scale 3D voxel and 2D BEV feature maps constructed from the input scan. By replacing point-neighborhood operators such as farthest point sampling and $k$-nearest neighbor search with voxel- and BEV-based feature extraction, our architecture is faster and can handle different input resolutions by design.
Experiments on SemanticKITTI and KITTI-360 show that our method achieves completion performance on par with the state of the art while completing a full scene in 0.1 seconds, which is 2.3 times faster than the fastest prior method. This matches the 10 Hz acquisition rate of typical automotive LiDAR sensors, taking a step toward real-time LiDAR scene completion. Code is available at \url{https://github.com/AzharSindhi/RapidLiDAR}.

\end{abstract}

\section{Introduction}
\label{sec:intro}

\begin{figure}[t]
    \centering
    \includegraphics[width=\linewidth]{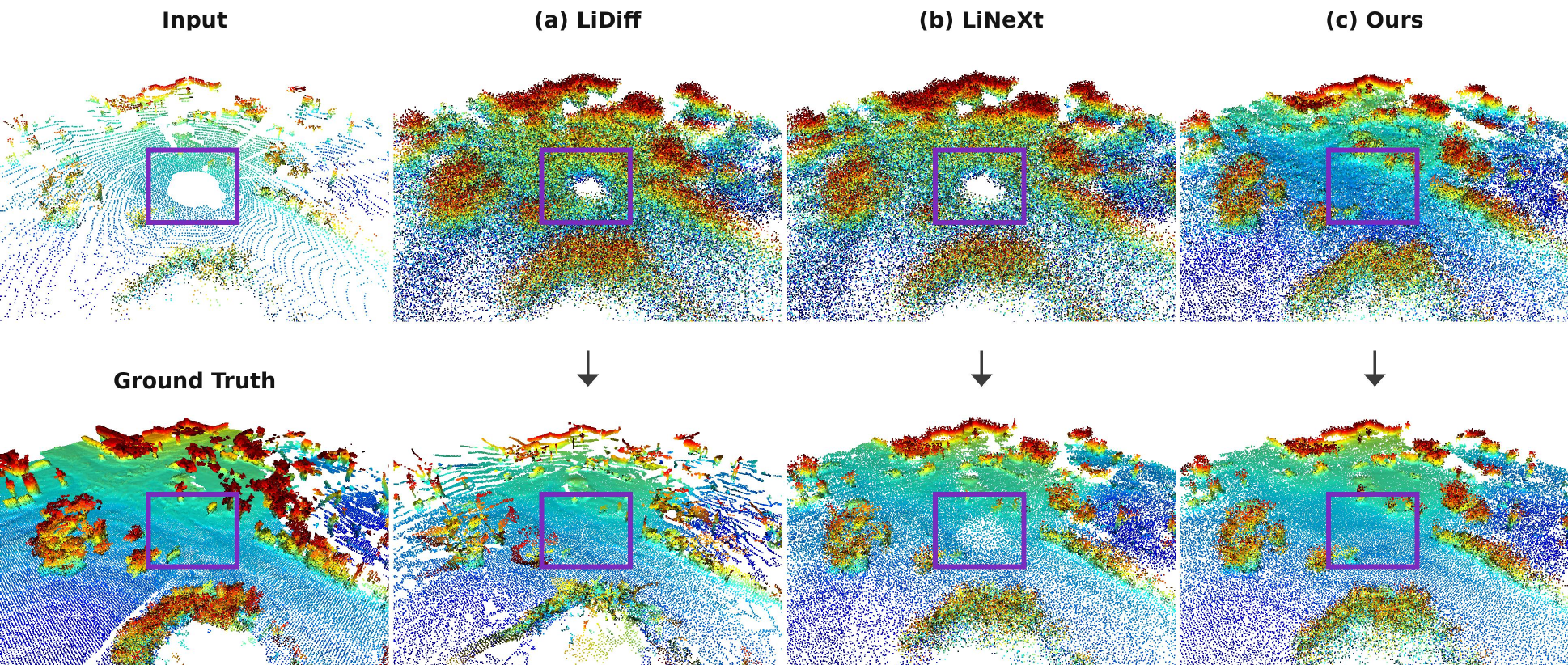}
    \caption{
    \textbf{Initialization matters.}
    \textit{Top row:} each method's initialization; \textit{bottom row:} the corresponding refined output. The highlighted box marks a large unobserved region. (a) LiDiff~\cite{nunes2024scaling} starts from Gaussian noise that carries
    no information about the scene; (b) LiNeXt~\cite{He2025LiNeXtRL} perturbs the input with a fixed noise variance, so its points stay near the observed surface and never reach across the gap; (c) Our adaptive module learns data-dependent displacements that populate the region from the surrounding geometry, and this coverage is preserved in both the coarse initialization and the final refined result.}
    \label{fig:teaser}
\end{figure}

LiDAR sensors are widely used in domains such as automated driving due to their ability to provide accurate 3D geometric measurements that are relatively more robust to lighting and weather conditions than cameras~\cite{Kim2023EmpiricalAO}. However, LiDAR measurements are often sparse and incomplete, particularly for distant objects where point density decreases, and in regions affected by occlusions or limited sensor resolution~\cite{Kim2023EmpiricalAO,li2023lode,vizzo2022make}. As a result, large portions of the environment may be only partially observed or entirely missing. This incomplete representation poses a significant challenge for downstream perception systems that rely on holistic scene understanding. Scene completion addresses this limitation by inferring missing geometry and reconstructing a dense, consistent representation of the surrounding environment.

Existing scene completion approaches~\cite{nunes2024scaling,martyniuk2025lidpm,liflow2025,He2025LiNeXtRL} generally follow an initialize-and-refine paradigm where an initial coarse estimate of the complete scene is first constructed, which we term the \textit{initialization}. This initialization is then refined into complete 3D geometry, depending on the specific refinement approach adopted in prior work. Dominant among these are generative models, particularly diffusion-based models~\cite{nunes2024scaling,martyniuk2025lidpm,liflow2025}, which use random Gaussian noise as the initialization and iteratively denoise it into a plausible 3D scene. On the other hand, a recent non-generative approach, LiNeXt~\cite{He2025LiNeXtRL}, initializes the scene by perturbing the partial input scene with fixed noise, which is then refined in a single forward pass.

We argue that the efficiency of scene completion methods, in terms of speed and generalization, is largely dependent on how well the initialization coarsely represents the final scene. For diffusion-based generative models, the random Gaussian noise used as initialization is least informative about the final complete scene; therefore, they require multiple iterations during inference to construct a plausible 3D scene. This iterative inference results in slower completion times, even with recent acceleration techniques~\cite{Zhang2024DistillingDM}. This makes generative models infeasible for time-critical applications such as automated driving, where real-time perception is essential. On the other hand, the fixed-noise-perturbed initialization strategy used by LiNeXt is not generalizable across different sensor configurations. Because sensor configurations vary in point density and coverage pattern, the noise scale must be retuned manually for each new dataset. Furthermore, because each point is perturbed only within a small, fixed radius of its original position, the initialized points remain close to the observed input and cannot cover large gaps and occluded regions.

In this work, we make the initialization itself a learned component. We propose an \textit{\textbf{adaptive initialization module}} that predicts a spatially varying displacement for each partial input point. Rather than perturbing points with a fixed noise scale, the module learns to displace each point by an amount that reflects the local scene structure: points near sparse or occluded regions move further to fill in the missing geometry, while points in already dense regions move only slightly. This produces a coarse scene initialization that is adapted to the local geometry, rather than uniformly spread around the observed input (see Fig.~\ref{fig:teaser}). To refine the adaptively initialized scene into a complete and coherent scene, we further propose a \textit{\textbf{multi-scale reconstruction module}} that adjusts point positions using multi-scale voxel and bird's-eye-view (BEV) features extracted from the input scan. Both modules extract features independently of the input point count by avoiding point-neighborhood operators, such as farthest point sampling (FPS) and $k$-nearest neighbor ($k$-NN), that are commonly used in point-based feature extractors such as LiNeXt. As a result, our approach is faster and can handle different input resolutions by design.

Experiments on SemanticKITTI~\cite{Behley2019SemanticKITTIAD} and KITTI-360~\cite{Liao2021KITTI360AN} show that our method achieves completion performance on par with the state of the art while completing a full scene in 0.1 s, which is $2.3\times$ faster than the fastest prior method. This matches the 10 Hz scan rate of typical automotive LiDAR sensors and provides a path toward real-time scene completion in autonomous driving scenarios. Our main contributions are summarized as follows:
\begin{itemize}
    \item We introduce an adaptive initialization module that constructs the initialized point set via spatially varying learned displacements, rather than relying on fixed, hand-tuned, and dataset-specific noise scales that require manual recalibration for each new sensor configuration.
    
    \item We design an architecture that replaces point-neighborhood operations (e.g., FPS and $k$-NN) with voxel- and BEV-based feature extraction, thereby improving runtime and supporting different input resolutions by design.
    
    \item We achieve completion performance on par with the state of the art on SemanticKITTI~\cite{Behley2019SemanticKITTIAD} and KITTI-360~\cite{Liao2021KITTI360AN}, and complete a full scene in 0.1 seconds, which is $2.3\times$ faster than the fastest prior method and matches the 10 Hz scanning rate of typical automotive LiDAR sensors.
\end{itemize}

\section{Related Work}
\label{sec:related_work}

We review generative and single-pass approaches to scene completion, as well as the BEV attention mechanisms on which our method builds.

\subsection{Generative Models for Scene Completion}
Following the success of Denoising Diffusion Probabilistic Models (DDPMs)~\cite{ho2020denoising} on synthetic single-object point clouds~\cite{luo2021diffusion,Zhou_2021_ICCV,lee2023diffusion}, generative models have become the dominant approach to real-world LiDAR scene completion~\cite{nunes2024scaling,martyniuk2025lidpm}. To preserve fine geometric detail at the scene scale, these methods formulate the diffusion process directly on 3D points, treating completion as a point-level denoising problem~\cite{nunes2024scaling} whose performance strongly depends on the choice of starting point~\cite{martyniuk2025lidpm}. Since iterative sampling requires hundreds of network evaluations and tens of seconds per scan, subsequent work has focused on reducing inference cost. Some methods distill a pre-trained diffusion model into a faster few-step student~\cite{Zhang2024DistillingDM}, while others replace denoising with flow matching to achieve competitive quality in fewer steps~\cite{liflow2025}. Despite these improvements, all generative approaches still require multiple network evaluations at inference time, which limits their use in real-time settings.

\subsection{Non-Generative Point Cloud Completion}
Single-pass completion has long been standard at the object level, where methods using synthetic benchmarks~\cite{yuan2018pcn,Chang2015ShapeNetAI} largely follow a common pattern: encode the partial cloud, decode a coarse point set, and refine it in a coarse-to-fine manner~\cite{yuan2018pcn,xiang2021snowflakenet,hussian2026i2pref}. Transformer-based variants extend this by generating structure-aware queries for the missing regions via FPS and $k$-NN grouping~\cite{Yu2021PoinTrDP}, following early transformers applied to point clouds~\cite{Engel2020PointT}. Recently, LiNeXt~\cite{He2025LiNeXtRL} brought the single-pass paradigm to the scene scale. It constructs an initial point set by replicating the observed points and perturbing them with fixed-variance Gaussian noise, then refines the entire set in a single forward pass. This eliminates the need for iterative sampling and achieves substantial reductions in inference time compared with diffusion-based methods. However, it inherits the FPS and $k$-NN neighborhood operators of object-level architectures, whose cost grows rapidly with the hundreds of thousands of points in outdoor scenes, and its fixed-noise initialization introduces a dataset-specific prior that confines spatial coverage to local neighborhoods around the observed points, leaving large occluded regions weakly constrained.

The scene-level feature aggregation has been addressed in a parallel line of work on autonomous driving perception. Deformable attention~\cite{zhu2021deformabledetr} evaluates features at a small number of learned sampling locations instead of attending densely over all positions, and, combined with Bird's-Eye-View representations~\cite{Li2022BEVFormerLB}, has become a standard tool across 3D detection~\cite{Liu_2024_CVPR,Xue2025CorrBEVM3,Zhang2023SABEVGS}, segmentation~\cite{Xu2022CoBEVTCB,Ge2023MetaBEVSS}, and occupancy prediction~\cite{Tong2023SceneAO,Li2024ViewFormerES,Agro2024UnOUO}. However, they operate on a small set of learned object or grid queries that predict labels at fixed locations. In contrast, for scene completion, the queries are the hundreds of thousands of output points themselves, whose 3D positions must first be initialized to cover unobserved regions and are then refined by the aggregated context.

Our method follows the single-pass paradigm but makes the initialization a learned, structure-conditioned component rather than a fixed random perturbation, and replaces point-neighborhood operators with deformable cross-attention over multi-scale BEV features, where the queries are the output points themselves, initialized to cover unobserved regions and refined by the aggregated context.

\section{Method}
\label{sec:method}

\begin{figure}
\centering
\includegraphics[width=0.8\linewidth]{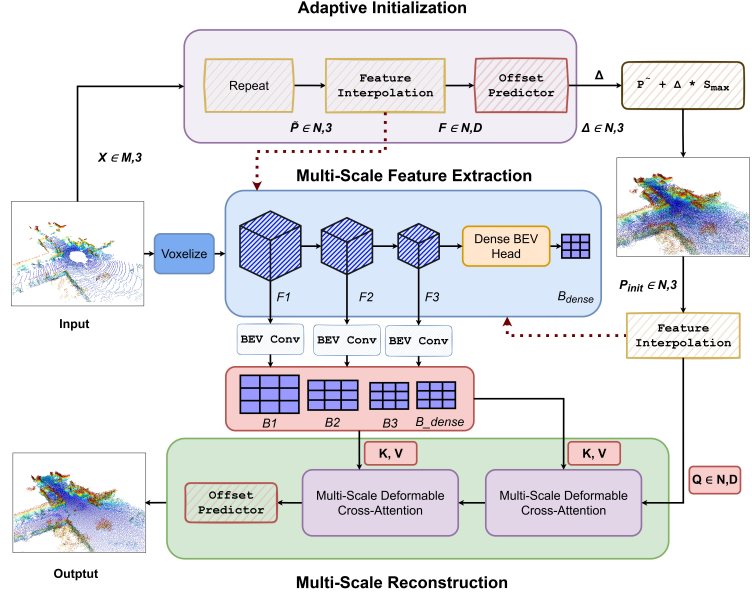}
\caption{\textbf{Overview of \methodname.} The \textbf{Multi-Scale Feature Extraction} module (\textit{center}) voxelizes the input point cloud $X \in \mathbb{R}^{M \times 3}$ and extracts multi-scale 3D voxel features $(F_1, F_2, \dots, F_n)$ and a dense 2D BEV feature map $B_{\text{dense}}$. The \textbf{Adaptive Initialization Module} (\textit{top}) predicts a spatially varying displacement $\Delta$ for each point in $\tilde{P}$ to obtain the initialized scene $P_{\text{init}}$. The \textbf{Multi-Scale Reconstruction Module} (\textit{bottom}) projects voxel features into BEV feature maps and cross-attends per-point features from $P_{\text{init}}$ with the multi-scale BEV feature maps using multi-scale deformable attention. A final MLP predicts a residual displacement that aligns each point with the underlying target surfaces, producing the completed scene.}
\label{fig:arch}
\end{figure}

Given an incomplete LiDAR point cloud $X \in \mathbb{R}^{M \times 3}$, the goal is to predict a complete scene $P \in \mathbb{R}^{N \times 3}$, where $M$ and $N$ denote the number of input and output points, respectively, with $N > M$. We obtain the complete scene in a single forward pass from $X$ to $P$ by training a neural network that directly optimizes the difference between the predicted and ground-truth scenes. Our overall architecture is illustrated in Fig.~\ref{fig:arch}.

We first voxelize the input $X$ and obtain multi-scale features (Sec.~\ref{sec:feature_extraction}). Using these multi-scale features, the scene is completed in two stages, both trained end-to-end. In the first stage, we propose an \emph{adaptive initialization module} that learns the initial coarse scene by predicting a spatially varying displacement for each point in $\tilde{P}$, which is an expanded version of the partial input scene $X$ (Sec.~\ref{sec:adaptive}). In the second stage, we propose a \emph{multi-scale reconstruction module} that refines this initial scene into a complete and coherent scene (Sec.~\ref{sec:refinement}). Finally, the network is trained end-to-end to minimize the Chamfer distance between the reconstructed and ground-truth scenes (Sec.~\ref{sec:loss}).

\subsection{Multi-Scale Feature Extraction}
\label{sec:feature_extraction}

\begin{figure}
\centering
\includegraphics[width=\linewidth]{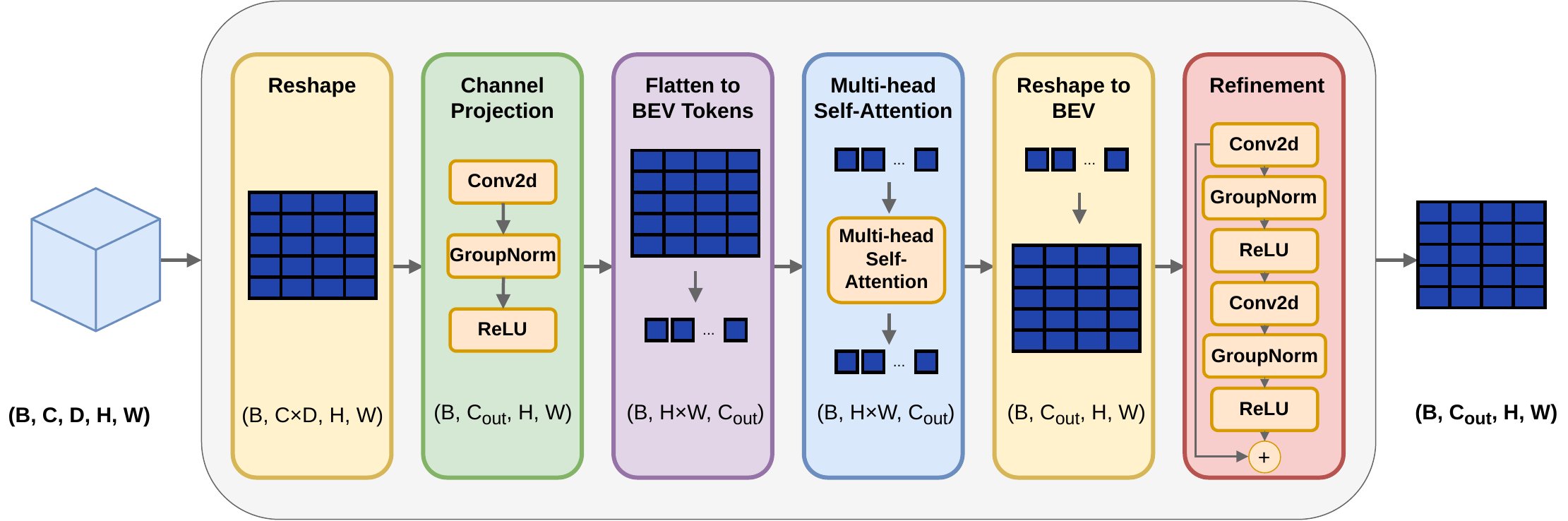}
\caption{\textbf{Illustration of Dense BEV Head.} Converts sparse 3D volumetric features into a dense BEV map. The channel and depth dimensions are merged and projected to $C_{\text{out}}$ using a 2D convolution. Multi-head self-attention over BEV tokens captures global scene context, followed by residual 2D convolutions for feature refinement.}
\label{fig:bev_head}
\end{figure}

In this module, we obtain multi-scale features from the partial input scene $X$. These features are then used in the subsequent stages to produce the initial coarse scene and the final completed scene.

To this end, we first voxelize the partial scene $X$ at a resolution of $\eta$, which results in a voxel grid of size $[1, D, H, W]$, where $D$, $H$, and $W$ denote the depth (Z-axis), height (Y-axis), and width (X-axis), respectively, and the single channel indicates whether the corresponding voxel is occupied. We then apply a sequence of 3D convolutional layers, each of which downsamples the input by a factor of 2, to obtain a set of multi-scale voxel features $\{F_i\}_{i=1}^{n}$, where $F_i \in \mathbb{R}^{C_i \times d_i \times h_i \times w_i}$, with $d_i = D/2^i$, $h_i = H/2^i$, and $w_i = W/2^i$ denoting the depth, height, and width at scale $i$.

While these voxel features capture fine-grained geometry, they remain sparse as only a small percentage of voxels are occupied, and the vast majority are empty due to the sparsity of LiDAR scans. Therefore, a dense representation of the scene is important and should also encode information about the surrounding sparse and occluded regions.

To obtain the dense feature map, we project the last voxel feature $F_n \in \mathbb{R}^{C_n \times d_n \times h_n \times w_n}$ into a 2D bird's-eye-view (BEV) map using a dedicated BEV head, which applies a series of operations. Specifically, we first reshape $F_n$ by merging its channel and depth dimensions into $(C_n \cdot d_n, h_n, w_n)$, and then project this representation to a BEV feature map $B_{\text{proj}} \in \mathbb{R}^{C_{\text{out}} \times h_n \times w_n}$ using a single 2D convolution.

To further enrich the features with information from neighboring regions, we apply multi-head self-attention (MHSA)~\cite{Vaswani2017AttentionIA}. Specifically, we treat each spatial location of $B_{\text{proj}}$ as an individual token, reshaping it into a sequence of tokens $S \in \mathbb{R}^{(h_n \cdot w_n) \times C_{\text{out}}}$, which is refined through MHSA. This enables the sequence $S$ to carry surrounding spatial context and encode information about sparse and occluded regions. Finally, we reshape the sequence $S$ back into a 2D feature map $(C_{\text{out}}, h_n, w_n)$ and further refine it with two convolutional layers connected by a skip connection. This produces the dense BEV feature map $B_{\text{dense}}$. We illustrate the BEV head operations in Fig.~\ref{fig:bev_head}.

Together, the sparse multi-scale 3D voxel features $F_1, F_2, \ldots, F_n$ and the dense 2D BEV feature map $B_{\text{dense}}$ form a hybrid representation of the scene. Both the \emph{adaptive initialization module} and the \emph{multi-scale reconstruction module} use this hybrid representation to construct the coarse initialized scene and the final completed scene, respectively.

\subsection{Adaptive Initialization Module}
\label{sec:adaptive}

Given the partial input scene $X \in \mathbb{R}^{M \times 3}$ and the multi-scale features $F_1, F_2, \ldots, F_n$ and $B_{\text{dense}}$, the objective of this module is to learn the coarse scene $P_{\text{init}} \in \mathbb{R}^{N \times 3}$ that serves as initialization for further refinement. Since the number of input points $M$ is smaller than the number of output points $N$ required to complete the scene, we repeat each observed point $(\lfloor N/M \rfloor)$ times and perturb the repeated points with small Gaussian noise $\sigma_{\text{init}}$ to avoid duplicates. Concatenating these perturbed points with the original partial scene $X$ gives the point set $\tilde{P} \in \mathbb{R}^{N \times 3}$ that matches the required number of target points.

To learn a coarse scene that can better cover sparse and occluded regions, we predict a spatially adaptive displacement for each point in $\tilde{P}$. By moving each point according to its learned displacement, the points are adaptively redistributed to match the target scene structure.

To predict a displacement for each point, we need a feature vector that captures both local geometry and broader scene context around that point. The multi-scale voxel features $F_1, F_2, \ldots, F_n$ and the dense BEV feature map $B_{\text{dense}}$ already contain this information, but are defined over their respective grids rather than over individual points. We therefore obtain a feature for each point by interpolating from these grids. Specifically, for each point $p_i = (x, y, z) \in \tilde{P}$, we project $p_i$ onto the voxel grid of each multi-scale feature map $F_k$, obtaining a voxel index $\sigma_k(p_i)$ at scale $k$, and sample the corresponding feature vector $f_i^{(k)} \in \mathbb{R}^{C_k}$ through interpolation. Similarly, we project $p_i$ onto the BEV grid to obtain a cell index $\pi(p_i)$ in $B_{\text{dense}}$ and sample the corresponding feature vector $b_i \in \mathbb{R}^{C_{\text{out}}}$. Concatenating these gives the per-point feature vector
\begin{equation}
f_i = \text{concat}\left(f_i^{(1)}, f_i^{(2)}, \ldots, f_i^{(n)}, b_i\right) \in \mathbb{R}^{d},
\end{equation}
where $d = C_1 + C_2 + \cdots + C_n + C_{\text{out}}$. Doing this for every point in $\tilde{P}$ in parallel gives the per-point feature matrix $\mathcal{F} \in \mathbb{R}^{N \times d}$.

Since directly indexing into these grids is a discrete, non-differentiable operation, we instead sample features using trilinear interpolation on the 3D voxel grids and bilinear interpolation on the 2D BEV grid, which allows smooth gradient flow during backpropagation. Trilinear interpolation distributes the sampling weights among the eight voxels neighboring $\sigma_k(p_i)$, while bilinear interpolation distributes the weights among the four cells neighboring $\pi(p_i)$.

Finally, the per-point feature matrix $\mathcal{F} \in \mathbb{R}^{N \times d}$ is passed through a Multi-Layer Perceptron (MLP) that predicts a spatially varying displacement $\Delta \in \mathbb{R}^{N \times 3}$. The coarse scene initialization $P_{\text{init}} \in \mathbb{R}^{N \times 3}$ is then obtained as
\begin{equation}
P_{\text{init}} = \tilde{P} + \Delta \cdot S_{\max},
\end{equation}
where $S_{\max}$ is a fixed scalar that defines the scale of the scene.

The learned displacement for each point can adapt to the local scene structure, in contrast to a fixed global noise. This adaptive strategy provides a better-informed initialization for the subsequent multi-scale reconstruction module, detailed in the following section.

\subsection{Multi-Scale Reconstruction Module}
\label{sec:refinement}

Given the coarse initialization $P_{\text{init}}$ from the \emph{adaptive initialization module}, and the multi-scale features $F_1, \dots, F_n, B_{\text{dense}}$ from the feature extraction module, the goal of this module is to reconstruct a complete and coherent 3D scene.

To this end, we obtain a feature vector for each point in $P_{\text{init}}$ through the same interpolation procedure described in the previous section, giving per-point features $\mathcal{F} \in \mathbb{R}^{N \times d}$. We reconstruct the scene using a transformer-based decoder, where the per-point features act as queries and the multi-scale features $F_1, \dots, F_n, B_{\text{dense}}$ act as context. A standard transformer decoder aggregates context through cross-attention~\cite{Vaswani2017AttentionIA}, in which each of the $N$ point queries attends to every position in $F_1, \dots, F_n, B_{\text{dense}}$. This scales as $\mathcal{O}(N \cdot K_{\text{ctx}})$, where $K_{\text{ctx}}$ is the total number of context elements across spatial grids. Given hundreds of thousands of point queries, full attention across all volumetric and BEV locations is computationally prohibitive.

Therefore, we use multi-scale deformable attention~\cite{zhu2021deformabledetr} between the point features $\mathcal{F}$ and the multi-scale features $F_1, \dots, F_n, B_{\text{dense}}$, which act as context. Deformable attention restricts attention to a small set of learned sampling locations within each context feature, rather than attending to all positions within each scale. This allows the point features to gather the relevant information required to complete the scene, without incurring high complexity.

However, current deformable attention implementations are optimized primarily for 2D feature maps rather than for 3D voxel features $F_1, F_2, \dots, F_n$. We therefore also project each of these features to 2D BEV representations. Specifically, given a voxel feature $F_i$ of shape $(C_i, d_i, h_i, w_i)$, we merge its channel and depth dimensions into $(C_i \cdot d_i, h_i, w_i)$ and project it to $(C_{\text{out}}, h_i, w_i)$ using a 2D convolution. This results in multi-scale BEV feature maps $B_1, \dots, B_n$, one for each 3D voxel feature.

We apply multi-scale deformable attention, with the point features $\mathcal{F}$ as queries and the multi-scale BEV features $\{B_1, B_2, \dots, B_n, B_{\text{dense}}\}$ as keys and values. This gives the refined features $F_{\text{ref}}$ as
\begin{equation}
F_{\text{ref}} = \text{MS-DeformAttn}\big(\mathcal{F}, \pi(P_{\text{init}}), \{B_1, \dots, B_n, B_{\text{dense}}\}\big),
\end{equation}
where $\pi(\cdot)$ projects the 3D coordinates in $P_{\text{init}}$ onto the BEV plane to obtain reference points for the attention mechanism, and $\{B_1, \dots, B_n, B_{\text{dense}}\}$ denotes the multi-scale BEV feature maps, including the dense BEV map from the BEV head.

Subsequently, the refined features $F_{\text{ref}}$ are passed through an MLP to predict a final 3D residual displacement $\Delta P_{\text{ref}} \in \mathbb{R}^{N \times 3}$. This residual allows the network to adjust the initialized points so that they can align with the underlying target surfaces. The final completed scene is obtained by applying the displacement:
\begin{equation}
P = P_{\text{init}} + \Delta P_{\text{ref}}.
\end{equation}

This refinement enables the network to use long-range geometric context to construct the scene without relying on point-neighborhood or spatial downsampling operations such as farthest point sampling (FPS) or $k$-nearest neighbor ($k$-NN), whose computational cost scales with the number of points.

\subsection{Loss Formulation}
\label{sec:loss}

Our network is trained end-to-end by directly optimizing the geometric fidelity of the predicted point sets. To quantify the spatial discrepancy between a predicted point cloud $P$ and the ground-truth complete scene $P_{\text{gt}}$, we use the standard Chamfer Distance (CD) metric. Chamfer Distance computes bi-directional nearest-neighbor distances between points in $P$ and $P_{\text{gt}}$, encouraging the prediction to cover the ground-truth surfaces while penalizing outliers:
\begin{equation}
\mathcal{L}_{\text{CD}}(P, P_{\text{gt}}) = \frac{1}{|P|} \sum_{p \in P} \min_{q \in P_{\text{gt}}} |p - q|_2^2 + \frac{1}{|P_{\text{gt}}|} \sum_{q \in P_{\text{gt}}} \min_{p \in P} |q - p|_2^2.
\label{eq:cd}
\end{equation}

\section{Experiments}
\label{sec:experiments}

In this section, we evaluate the effectiveness, efficiency, and generalizability of our method. We show that our proposed method achieves state-of-the-art performance while reducing inference time. We also assess the zero-shot behavior of our model without additional retraining.

\subsection{Datasets and Evaluation Metrics}
\label{sec:datasets}

\paragraph{Datasets.}
Following the standard protocols established by recent scene completion benchmarks~\cite{nunes2024scaling,Zhang2024DistillingDM}, we train and evaluate our method on the SemanticKITTI~\cite{Behley2019SemanticKITTIAD} dataset. Specifically, we use sequences 00--10 for training, excluding sequence 08, which is reserved for validation. To test the zero-shot generalization, we directly evaluate the model trained on SemanticKITTI on sequence 00 of the KITTI-360~\cite{Liao2021KITTI360AN} dataset without fine-tuning. During training, the partial scene is downsampled to 18{,}000 points, and the corresponding complete scene contains 180{,}000 points, following the standard practice. 

\paragraph{Evaluation Metrics.}
To evaluate completion performance, we use standard geometric metrics. We measure overall geometric accuracy using the Chamfer Distance (CD), computed as a bidirectional nearest-neighbor distance, as defined in Eq.~\ref{eq:cd}. To assess similarity in spatial distribution, we compute Jensen--Shannon divergence (JSD) in both 3D space and Bird's-Eye-View (BEV) representation.

\subsection{Implementation Details}
\label{sec:implementation}

Following prior benchmarks, we set $M = 18{,}000$ and $N = 180{,}000$, which yields an expansion ratio $\lfloor N/M \rfloor = 10$. The input is voxelized at a resolution of $\eta = 0.3\,$m to form an occupancy grid of size $[1, D, H, W] = [1, 20, 333, 333]$, which the multi-scale feature extraction backbone processes at $n = 4$ levels with channel dimensions $[C_1, C_2, C_3, C_4] = [32, 64, 128, 256]$. The last level is compressed by the BEV head into $C_{\text{out}} = 512$ channels with $4$ self-attention heads. We set the noise scale $\sigma_{\text{init}} = 0.1\,$m for the initial point repetition. During feature interpolation, the sampled features from all four voxel levels and the dense BEV map are concatenated into a per-point feature of dimension $d = C_1 + C_2 + C_3 + C_4 + C_{\text{out}} = 992$, then projected to the model width $C_{\text{out}} = 512$. The predicted displacement is bounded by a maximum expansion $S_{\max} = 50\,$m. In the reconstruction module, each voxel level is likewise projected to a BEV map with $C_{\text{out}} = 512$ channels, for a total of $L = 5$ feature levels $\{B_1, \dots, B_4, B_{\text{dense}}\}$ that serve as keys and values for $2$ deformable cross-attention stages with $n_h = 8$ heads and $K = 4$ sampling points per head.

We train with the Adam optimizer, a base learning rate of $1 \times 10^{-4}$, and a cosine learning rate schedule, using a batch size of $4$ on a single NVIDIA RTX~6000~Ada GPU; the model converges within $30$ epochs. Inference times are measured on the same GPU for all methods, using official code and released checkpoints.

\paragraph{Refinement Network.} Existing baselines additionally train a refinement network on top of the base completion model, first introduced by Nunes et al.~\cite{nunes2024scaling}. To ensure a fair comparison, we likewise train a refinement network on top of our completion model. Our refinement network reuses the multi-scale features extracted by the frozen completion network: per-point features are interpolated at each completed point, and four successive stages of deformable cross-attention over the BEV maps predict $\kappa = 6$ residual offsets per point, upsampling the scene by a factor of $6\times$. The refinement network is trained separately for 5 epochs, optimizing the Chamfer Distance in Eq.~\ref{eq:cd}.

\subsection{Scene Completion}
\label{sec:main_results}

Table~\ref{tab:main_comparison} summarizes quantitative results on SemanticKITTI~\cite{Behley2019SemanticKITTIAD} and KITTI-360~\cite{Liao2021KITTI360AN}. On SemanticKITTI, our method achieves the best results across all reported metrics, attaining the lowest Chamfer Distance, 3D JSD, and BEV JSD among all compared methods. Similarly, on KITTI-360, evaluated in a cross-dataset setting without fine-tuning, our approach follows a similar trend and achieves state-of-the-art performance.

\begin{table}[t]
    \centering
    \caption{\textbf{Scene Completion on SemanticKITTI and KITTI-360.} Quantitative comparison with prior methods. $\dagger$ denotes methods with an additional refinement. Best results in each group are highlighted in \textbf{bold}.}
    \label{tab:main_comparison}
    \setlength{\tabcolsep}{3.8pt}
    \begin{tabular}{@{}l ccc ccc@{}}
    \toprule
    & \multicolumn{3}{c}{SemanticKITTI} & \multicolumn{3}{c}{KITTI-360} \\
    \cmidrule(lr){2-4} \cmidrule(l){5-7}
    Method & CD $\downarrow$ & JSD 3D $\downarrow$ & JSD BEV $\downarrow$ & CD $\downarrow$ & JSD 3D $\downarrow$ & JSD BEV $\downarrow$ \\
    \midrule
    LMSCNet    & 0.641 & --    & 0.431 & 0.979 & --    & 0.496 \\
    LODE       & 1.029 & --    & 0.451 & 1.565 & --    & 0.483 \\
    MID        & 0.503 & --    & 0.470 & 0.637 & --    & 0.476 \\
    PVD        & 1.256 & --    & 0.498 & --    & --    & --    \\
    LiDiff     & 0.434 & 0.564 & 0.444 & 0.564 & --    & 0.459 \\
    LiDPM      & 0.446 & 0.532 & 0.440 & --    & --    & --    \\
    ScoreLiDAR & 0.406 & --    & 0.425 & 0.472 & --    & 0.444 \\
    LiFlow     & 0.309 & --    & 0.416 & --    & --    & --    \\
    LiNeXt     & 0.214 & 0.494 & 0.336 & 0.217 & 0.508 & 0.355 \\
    \textbf{Ours} & \textbf{0.206} & \textbf{0.475} & \textbf{0.332} & \textbf{0.211} & \textbf{0.492} & \textbf{0.338} \\
    \midrule
    LiDiff$^\dagger$     & 0.376 & 0.573 & 0.416 & 0.517 & --    & 0.446 \\
    ScoreLiDAR$^\dagger$ & 0.342 & --    & 0.399 & 0.452 & --    & 0.437 \\
    LiDPM$^\dagger$      & 0.376 & 0.542 & 0.403 & --    & --    & --    \\
    LiNeXt$^\dagger$     & 0.149 & 0.481 & 0.331 & 0.149 & 0.499 & 0.339 \\
    Ours$^\dagger$       & \textbf{0.138}    & \textbf{0.478}    & \textbf{0.330}    & \textbf{0.140}    & \textbf{0.490}    & \textbf{0.336}    \\
    \bottomrule
    \end{tabular}
\end{table}

\begin{table}
    \centering
    \caption{\textbf{Computational Efficiency Comparison.} We report Chamfer distance, number of learnable parameters, and inference time per scan.}
    \label{tab:efficiency_comparison}
    \setlength{\tabcolsep}{8pt} 
    \renewcommand{\arraystretch}{1.15} 
    \begin{tabular}{@{}l c c c@{}}
    \toprule
    Method & CD $\downarrow$ & Param (M) $\downarrow$ & Time (s) $\downarrow$ \\
    \midrule
    LiDiff & 0.434 & 32.67 & 30.1 \\
    ScoreLiDAR & 0.406 & 32.67 & 7.1 \\
    LiNeXt & 0.214 & 1.99 & 0.23 \\
    \textbf{Ours} & \textbf{0.206} & 11.8 & \textbf{0.10} \\
    \bottomrule
    \end{tabular}
\end{table}

Our main advantage over existing methods is speed. Table~\ref{tab:efficiency_comparison} reports the computational cost of different approaches. Diffusion-based generative models, such as LiDiff and ScoreLiDAR, require substantially more parameters and incur longer inference times due to their multi-step sampling. LiNeXt uses a compact single-pass architecture with a small parameter count but runs at around 0.23\,s per scan, despite its point-neighborhood operations relying on efficient custom CUDA implementations. Our method uses a moderate number of parameters and completes a full scene in 0.1\,s, which is $2.3\times$ faster than LiNeXt while also achieving a lower Chamfer Distance. This matches the 10\,Hz acquisition rate of typical automotive LiDAR sensors. Notably, our implementation relies only on standard library operations rather than specialized CUDA kernels. The accuracy and efficiency results demonstrate that our approach achieves the best trade-off among prior methods for reconstruction quality, model size, and inference speed.

\subsection{Ablation Studies}
\label{sec:ablations}

We perform ablation studies to analyze the influence of the main architectural components and voxel resolution on reconstruction quality and efficiency. First, to assess the contribution of each component, we perform an ablation study on the SemanticKITTI validation set, summarized in Table~\ref{tab:ablation_architecture}. In the first ablation, we replace the Adaptive Initialization Module (AIM) with a fixed perturbation scale of $\sigma_{\text{init}}=1.0$ following LiNeXt~\cite{He2025LiNeXtRL}, removing per-point displacement prediction. In the second, we remove the Multi-Scale Reconstruction Module (MSRM) and predict the final scene directly from the adaptively initialized point set, bypassing deformable cross-attention entirely. Both ablations lead to a consistent drop across all three metrics, indicating that each component contributes to the final result.

\begin{table}
\centering
\caption{\textbf{Architectural Ablation Study.} We evaluate the impact of our core modules on the SemanticKITTI validation set. Best results are in \textbf{bold}.}
\label{tab:ablation_architecture}
\begin{tabular}{l  c c c}
\toprule
Method & CD $\downarrow$ & JSD 3D $\downarrow$ & JSD BEV $\downarrow$\\
\midrule
Ours & \textbf{0.206} & \textbf{0.475} & \textbf{0.332}  \\
\midrule
Ours w/o AIM & 0.218 & 0.488 & 0.345 \\
Ours w/o MSRM & 0.215 & 0.494 & 0.342 \\
\bottomrule
\end{tabular}
\end{table}

Furthermore, we evaluate the effect of the maximum displacement bound $S_{\max}$ on reconstruction performance. We train the network with $S_{\max} \in \{50, 70, 100\}$ and report Chamfer Distance on a downsampled validation set of 180{,}000 output points. As shown in Table~\ref{tab:ablation_smax}, performance is largely insensitive to the choice of $S_{\max}$. This indicates that the adaptive initialization module is robust across a reasonable range of displacement scales, and we use $S_{\max} = 50$ as our default setting.

\begin{table}[h]
    \centering
    \caption{\textbf{Effect of Maximum Displacement Bound.} Reconstruction performance for different values of $S_{\max}$, evaluated on a downsampled validation set of 180{,}000 output points.}
    \label{tab:ablation_smax}
    \setlength{\tabcolsep}{10pt}
    \renewcommand{\arraystretch}{1.15}
    \begin{tabular}{@{}cc@{}}
    \toprule
    $S_{\max}$ & CD $\downarrow$ \\
    \midrule
    50  & 0.2594 \\
    70  & 0.2592 \\
    100 & 0.2589 \\
    \bottomrule
    \end{tabular}
\end{table}

\begin{table}
    \centering
    \caption{\textbf{Voxel Resolution Ablation.} Impact of voxel resolution $\eta$ on reconstruction performance, parameter count, and inference time.}
    \label{tab:resolution_comparison}
    \setlength{\tabcolsep}{8pt}
    \renewcommand{\arraystretch}{1.15} 
    \begin{tabular}{@{}l c c c@{}}
    \toprule
    $\eta$ (m) & CD $\downarrow$ & Param (M) & Time (s) $\downarrow$ \\
    \midrule
    0.5 & 0.214 & 10.0 & 0.07 \\
    0.4 & 0.210 & 11.6 & 0.09 \\
    0.3 & 0.206 & 11.8 & 0.10 \\
    0.2 & 0.208 & 11.9 & 0.14 \\
    \bottomrule
    \end{tabular}
\end{table}

Finally, Table~\ref{tab:resolution_comparison} reports the effect of voxel resolution on reconstruction quality, parameter count, and inference time. Finer resolutions consistently improve the Chamfer Distance, while the growth in both parameter count and inference time remains moderate. We select $\eta=0.3$ as it offers a suitable trade-off across all three criteria, though the architecture allows adjustment of voxel resolution to meet different speed or quality requirements.

\subsection{Qualitative Analysis}

\begin{figure}[t]
    \centering
    \includegraphics[width=0.6\linewidth]{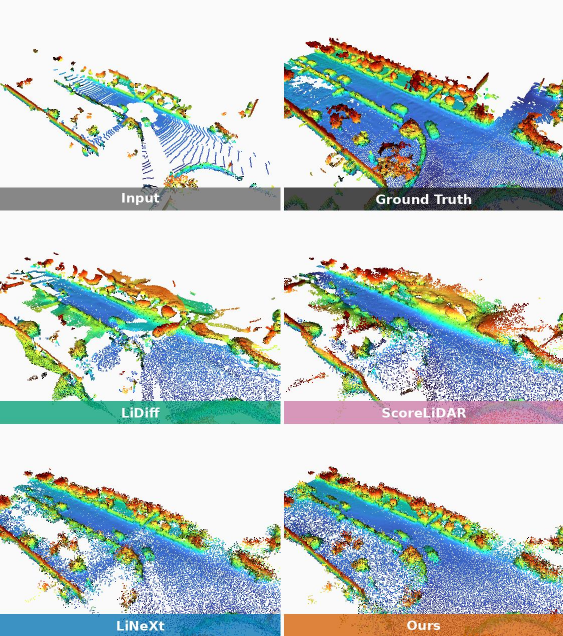}
    \caption{
    \textbf{Qualitative Comparison on SemanticKITTI.} Our method produces more complete geometry in large occluded regions compared to prior methods.}
    \label{fig:qual_semantickitti}
\end{figure}

We also examine qualitative results on SemanticKITTI to visualize how different methods reconstruct occluded regions and large-scale scene geometry. Representative examples for SemanticKITTI are shown in Figure~\ref{fig:qual_semantickitti}. Generative diffusion-based methods such as LiDiff and ScoreLiDAR produce visually plausible completions, but the reconstructed geometry can deviate from the ground-truth layout, especially in large occluded regions. LiNeXt, which directly optimizes the Chamfer Distance, often generates completions that are closer to the ground-truth but sometimes leaves parts of the scene under-completed due to its fixed-noise initialization. Our method produces completions that follow the ground-truth structure while providing broader coverage in regions with limited observations, which we attribute to the adaptive initialization not being restricted around the observed points.


\section{Conclusion}

We introduced RapidLiDAR, a LiDAR scene completion method that learns the initialization itself directly from data. Its adaptive initialization module predicts a spatially varying displacement for each partial input point, expanding these observations into a coarse scene initialization that adapts to the local geometry and removes the need for manual noise tuning. A subsequent multi-scale reconstruction module then refines this coarse initialization into a complete and coherent scene by predicting residual displacements from multi-scale voxel and bird's-eye-view features extracted from the partial input scan. Because it replaces point-neighborhood operators, such as farthest point sampling and $k$-nearest neighbor search, with voxel- and BEV-based feature extraction, the resulting architecture runs faster and handles different input resolutions.

On SemanticKITTI and KITTI-360, our experiments confirmed completion performance on par with the state of the art, with a full scene completed in 0.1 seconds, which is 2.3 times faster than the fastest prior method and in line with the 10 Hz acquisition rate of typical automotive LiDAR sensors.

We plan to extend this work to generalize across different sensor configurations. Evaluating robustness against domain shifts in sensor beam geometries offers a promising direction for future research.

\section*{Acknowledgment}
We acknowledge funding received from the Bavarian Ministry of Economic Affairs, Regional Development and Energy, in the scope of the funded project ”Bavarian Advanced Resolution Radar (BAVAR-RADAR)”, funding label DIK0622.

%
%
\bibliographystyle{splncs04}
\bibliography{main}

\end{document}